# Utilizing Post-Hurricane Satellite Imagery to Identify Flooding Damage with Convolutional Neural Networks

Jimmy Bao[1], Patrick Emedom-Nnamdi

## Abstract

Post-hurricane damage assessment is crucial towards managing resource allocations and executing an effective response. Traditionally, this evaluation is performed through field reconnaissance, which is slow, hazardous, and arduous. Instead, in this paper we furthered the idea of implementing deep learning through convolutional neural networks in order to classify post-hurricane satellite imagery of buildings as 'Flooded/Damaged' or 'Undamaged'. The experimentation was conducted employing a dataset containing post-hurricane satellite imagery from the Greater Houston area after Hurricane Harvey in 2017. This paper implemented three convolutional neural network model architectures paired with additional model considerations in order to achieve high accuracies (over 99%), reinforcing the effective use of machine learning in post-hurricane disaster assessment.

## 1. Introduction

Natural disasters, such as hurricanes, are extreme meteorological, hydrological or geological events that cause tremendous damage to affected communities. There are three crucial focal points for disaster assessment: distinguish which parts of the community have been disproportionately affected, determine if external help is required, and develop projections to create predictions for future natural disasters [1,7].

---

[1] Corresponding author: Jimmy Bao
UT Austin, Austin, Texas, USA
Email: jimmybao22@utexas.edu

In addition to heavy wind and rain, hurricanes are known to cause flooding and bring tremendous damage to coastal communities. In the 20th century, the U.S. Geological Survey (USGS) discovered that floods were the most dangerous natural disasters in the world in terms of casualties [2]. Among these natural disasters was Hurricane Katrina, a category 5 hurricane at its peak. Hurricane Katrina caused unprecedented damage to Mississippi and Louisiana, claiming over 1,800 lives and causing approximately $125 billion damage [3].

Traditionally, hurricane damage has been assessed through field reconnaissance, where volunteers and workers record all damage-related information. Nevertheless, in recent years, satellite and aerial imagery have become increasingly accessible [4]. For example, synthetic-aperture radar (SAR) imagery has been utilized in flood detection [5]. The SAR images are helpful in mapping features, texture, and roughness patterns. Although these images are effective in assessing damage at the region-level, the quality is not yet fit to assess damage at the building-level [6]. Thus, in this paper, we instead recommend optical sensor imagery as a more practical and accessible method for analyzing hurricane damage at the building-level. However, processes that rely on direct human classification of optical sensor images can be both laborious and prone to human error. Thus, machine learning based approaches to image classification pose a very appealing solution that removes the potential for human error. In this paper, we will utilize machine learning to label images obtained from the optical sensors as 'Flooded/Damaged' or 'Undamaged'. This will reduce the response time to the disaster and provide an accurate indication as to which parts of the community require the most support. To achieve this, we introduce two convolutional neural networks (detailed in section 3.3.2) that are trained on a dataset (detailed in section 3.1) containing labeled images of hurricane disasters.

The remainder of this paper is organized as follows. Section 2 will describe recent works that have been made in this field. Section 3 will introduce our methodology, including a brief description of the dataset, a high level analysis of convolutional neural networks, and the pipeline for the models. Section 4 will present the results achieved by the models. Finally, Section 5 will conclude the paper and discuss the significance of our findings.

## 2. Related Works

Implementing computer vision techniques to detect natural weather conditions and assess damage has been a prominent topic recently. In many cases, studies have been performed to create models on weather classification, that distinguish between snow, rain, fog, and normal weather [8]; to detect natural disasters, damage, and incidents through social media images [9]; and to quickly determine building damage through satellite images [14]. Heading to more specific instances, models have also been created to assess a wide array of natural disasters, such as tornadoes [10,11], volcanoes [12,13], and earthquakes [15].

In particular, a lot of work has already been done in assessing floods. For example, Rahnemoonfar et al. created a dataset *FloodNet* based on aerial imagery and implemented models for classification, semantic segmentation, and visual question answering (VQA) of flooded areas. For classification, they trained three models: InceptionNetv3, ResNet50, and Xception, achieving accuracy as high as 95% [16]. Meanwhile, Gan and Zailah created a model to classify water level and risks, achieving upwards of 99% accuracy [17]. For works related to hurricane damage assessment, Wang and Qu created an algorithm to assess post-hurricane forest damage [18]. Meanwhile, Barnes et al. utilized satellite images after Hurricane Katrina for damage assessment and emergency response planning [20]. Furthermore, Khar et al. compiled numerous models built by researchers used for detecting damage caused by natural disasters, especially hurricane damage. This included a model on damage-assessment post-Hurricane Sandy that achieved 88.3% accuracy [19]. Additionally, Cao and Choe created a model post-Hurricane Harvey based on an architecture they designed from satellite images, which achieved approximately 97% accuracy [6].

This paper will add on to the experimentations done by Cao and Choe. We will be implementing two popular convolutional neural network models (detailed in section 3.3.2) that attempt to discover a more reliable data-driven model. In the end, we will compare the convolutional neural network models to each other and to Cao and Choe's model.

## 3. Methodology

3.1 Dataset

The dataset was compiled by Cao and Choe, and is composed of satellite imagery captured by optical sensors. The images capture the Greater Houston area in the aftermath of Hurricane Harvey (see Figure 1). While the original raw images contained roughly 4,000 image strips, with each strip at around 1GB with 400 million pixels, Cao and Choe have already preprocessed the images for easier analysis [6]. In the end, each 128x128 pixelated image is placed in folders labeled as either 'damage' or 'no_damage' (see Figure 2). Then, the resulting dataset is pre-grouped into testing (1,000 images of each class), training (5,000 images of each class), and validation (1,000 images of each class) sets named 'test', 'train_another', and 'validation_another' respectively. Additionally, Cao and Choe have provided a dataset named

'test_another' which is unbalanced and contains 8,000/1,000 images of damaged/undamaged classes [22].

In creating this dataset, Cao and Choe hoped to produce a model that efficiently assesses post-hurricane damage. Cao and Choe have shared their dataset with the intention of motivating others to build on their research, allowing for a more efficient damage assessment process [6]. Hence, this paper will explore how recent popular convolutional neural network architectures perform on this dataset, and compare the findings to Cao and Choe's model.

3.2 Convolutional Neural Network

Before trying to grasp convolutional neural networks, one should first have a basic understanding of neural networks. Neural networks (NNs) were first introduced by McCulloch and Pitts in 1943 [26], and are viewed as simplified models of the neural processing that occurs in human brains. More specifically, they are composed of artificial neurons linked together in networks, and thus are also known as artificial neural networks (ANN) (see Figure 3). Neural networks are utilized to make data-driven predictions/decisions [29]. Over the years, additional improvements have been introduced into the field: backpropagation [28]; splitting the data into training, validation, and test sets [30]; cross-validation [31]; regularization [32]; stochastic gradient descent (SGD) [33]; batch normalization [34]; data augmentation [35]. The neural networks which contain multiple processing layers give name to a new field within artificial intelligence: deep learning [36]. The structure consists of the first layer (input layer), the last layer (output layer), and the layer(s) in between (hidden layer(s)) [32]. Most neural networks are also known as feedforward neural networks because information only travels forwards through the network [27]. Neural networks which contain feedback connections are called recurrent neural networks. The neural networks (including convolutional neural networks) described in this paper are feedforward neural networks. Each neuron within the neural network contains a set of learnable parameters: a bias value and a collection of weights. Each input is then multiplied by a weight, and the weighted sum of the inputs is added to the bias value. Finally, this is fed into a chosen activation function, giving the final output of the neuron. Neural networks apply gradient descent in order to determine the set of parameter values that minimize a chosen loss function. Through this method, neural networks seek to obtain a minimum loss [32].

Convolutional neural networks (CNNs) are a subset of artificial neural networks, and are most commonly applied in analyzing visual imagery [37,38] (see Figure 4). Convolutional neural networks contain three different components: convolutional layer(s), pooling layer(s), and fully connected layer(s). Convolutional layers are the center pieces of CNNs. These layers contain a set of learnable filters, each of which have small widths and heights, but extend through the whole depth of the input volume. The filters repeatedly take dot products of weighted parameters. For example, a 3x3 convolutional filter will take the dot product of 9 weighted parameters in a 3x3 block. This is then pushed through an activation function that later acts as an input to the next layer [21,32,46]. Common activation functions include sigmoid, rectified linear unit (ReLU), and leaky ReLU. For this paper, we will utilize the ReLU activation function as it has the potential to speed up the model training while maintaining consistent performance [45]. The pooling layer, also known as a subsampling layer, is utilized to reduce the dimensions of the inputs. Therefore, it is useful in reducing the number of parameters and computation required. It utilizes a small net and replaces the output value in each net by a new value (see Figure 5). The most popular, and the one implemented in this paper, is max pooling. This replaces the output value by the maximum within the net [46,47]. Finally, the network is flattened into a sequence of fully connected layers. The neurons in a fully connected layer (also known as the dense layer) are linked to all activations in the previous layer, like in a regular neural network. The output is determined in the same way as a normal neural network (inputs multiplied by weights, added onto a bias, and transformed through an activation function) [21]. For classification tasks, the softmax activation function is commonly employed in the last layer. This ensures that the values of each neuron corresponds to the probabilities of a single class for each image [32].

Overall, convolutional neural networks have excelled in image classification throughout numerous datasets, propelling its popularity. These include but are not limited to ImageNet [23], CIFAR-10 [40], MNIST [41], FashionMNIST [42], and The Street View House Numbers (SVHN) [43], which have all reached extraordinary accuracies. Thus, for this project, we will also employ convolutional neural networks.

## 3.3 Pipeline

### 3.3.1 Data Preprocessing

Since the dataset contains folders with jpeg satellite images, we utilized python to load the data into multidimensional numpy arrays. Since each image is 128x128 pixels, and each pixel has RGB values, the final multidimensional numpy array will have size Nx128x128x3, where N is the number of images.

To preprocess the data, we normalize the data by scaling the pixels. Since each RGB value is between 0 and 255, we divide each pixel by 255, which causes each RGB value to be between 0 and 1. We normalized the data, because it is necessary for obtaining adequate results and quickening the runtime of the neural network, according to Sola and Sevilla [21].

3.3.2 Model Architecture

We opted to train pre-existing model architectures which have performed exceptionally well on other image datasets rather than create our own. The three model architectures implemented in this paper are based on AlexNet [23] and VGGNet [39]. AlexNet is a convolutional neural network architecture that was designed by Krizhevsky et al., and was utilized in the ImageNet Large Scale Visual Recognition Challenge (ILSVRC) in 2012 (see Figure 6). Here, Krizhevsky et al. achieved a top-5 test error[2] of 15.3%, which was 10.9% lower than the second best model. Krizhevsky et al. highlighted the importance of the depth of his model, stating that removing a single layer would cause notable differences. For example, removing a middle layer could increase the top-1 test error[3] by 2%. However, because of the depth, the CNN was computationally expensive and was made feasible by the use of graphic processing units (GPUs) [23]. On the other hand, VGG is a convolutional neural network architecture that was designed by Simonyan and Zisserman in the Visual Geometry Group and was utilized in the ImageNet Large Scale Visual Recognition Challenge (ILSVRC) in 2014 (see Figure 7). The model was one of the top performers, achieving a top-5 test error of only 7.0%. Like AlexNet, Simonyan and Zisserman also emphasized the importance of the depth of the model [39].

There exists key differences between the architectures of the two models chosen. VGGNet employs much smaller receptive fields and stride (3x3 with a stride of 1) compared to AlexNet (11x11 with a stride of 4). Furthermore, VGGNet contains groups of 2 or 3 convolutional layers before every max pooling layer. Meanwhile, the first two convolutional layers in AlexNet are followed directly by a max pooling layer. Most crucially, VGGNet has much greater depth than AlexNet, and thus achieved a higher accuracy in ILSVRC [23,39].

---

[2] Top-5 test error is the percent of images in the test set where the true label does not match the model's top 5 predicted labels.
[3] Top-1 test error is the percent of images in the test set where the true label does not match the model's top predicted label.

Given the limited access to graphic processing units (GPUs), training time became one of the most important factors in determining which models to implement. We decided to employ the AlexNet architecture directly as presented by Krizhevsky et al. because of its relatively fast training time, its ease of implementation, and also its overall high accuracy in image classification. Similar to Krizhevsky et al., we also implemented dropout layers after the first two fully-connected layers, because otherwise Krizhevsky et al. warns that the model substantially overfits [23]. For VGGNet, we implemented and trained a VGG16 network, same as Simonyan and Zisserman's model D architecture in Figure 7. However, since this architecture takes an extensive amount of time to train without access to a GPU, we also implemented a 3-block VGG-style architecture. The 3-block VGGNet maintains the general structure of a VGGNet architecture (two convolutional layers followed by a max pooling layer), but instead of containing 16 or 19 weight layers, the 3-block VGGNet only has 9 weight layers (see Table 1) [39].

### 3.3.3 Model Training

In order to train our model with data, we employed the *Keras* library with TensorFlow [56] backend. We set the default mini batch size to 64. For all three model architectures we set the default learning rate to 0.001. Furthermore, the weights for the CNN were initialized using the He Uniform initializer [55]. The activation function utilized for the layers excluding the output layer is the rectified linear unit (ReLU) [45]. As aforementioned, the output layer employs the softmax activation function. The loss function employed is the cross-entropy loss. In fact, Weber et al. discovered that the most successful and reliable strategy for image classification is to implement a cross-entropy loss on top of a softmax classification function for the output layer [9]. We implemented the stochastic gradient descent (SGD) optimizer [33], with the default momentum set to 0.9. Because of the expensive computational cost of convolutional neural networks, we will tune the hyperparameters[4] greedily rather than in a full grid search. For example, if we determine that the best learning rate for a model is 0.001, we will utilize that learning rate when tuning other hyperparameters as well. Further, we will also choose which models to tune greedily, based on the few that performed the best with default parameters.

### 3.3.4 Model Evaluation

After the model completes training and tuning, the models will be compared to each other, and to Cao and Choe's model [6]. In addition to comparing accuracies, other measures will also be examined. More specifically we will focus on the confusion matrix or error matrix, which allows us to visualize the performance of the model [48]. For binary classification under the trained CNNs, we utilized a 2x2 matrix. The confusion matrix is found by predicting the labels of the test dataset with the trained CNN model, and then comparing these to the true labels. The confusion matrix is composed of 4 values (see Figure 8): *true negative*, when the instance is negative and classified as negative (i.e. the model correctly classified the image as undamaged); *false positive,* when the instance is negative but classified as positive (i.e. the model correctly classified the image as damaged); *false negative*, when the instance is positive but classified as negative (i.e. the model incorrectly classified the image as undamaged); and *true positive*, when the instance is positive and classified as positive (i.e. the model incorrectly classified the image as damaged). The numbers along the major diagonal represent the cases where the correct classification was made. The components of the confusion matrix can also be employed to calculate other valuable information. These include but are not limited to: true positive rate (TPR), also known as recall and hit rate, which equals the true positives over all actual positives; true negative rate (TNR), also known as specificity, which equals the true negatives over all actual negatives; positive predictive value (PPV), also known as precision, which equals the true positives over all predicted positives; negative predictive value (NPV) which equals the true negatives over all predicted negatives; and F-score, also known as F-measure or F1 score, which is the harmonic mean of recall (TPR) and precision (PPV) [49]. In essence, these values are different types of accuracy measurements , and selecting an appropriate one depends on the objective of the experimentation [48]. For this paper, we will compare accuracies as a base level of measurement. Another adequate unit of measurement to utilize would be one which allows us to minimize the *false negatives* obtained (e.g. recall, F1 score, etc.), since incorrectly classifying an image as undamaged could be detrimental. In section 4, we will calculate these values for each model.

---

[4] Hyperparameters are defined in section 3.5.

3.4 Validation

The validation set is obtained from dividing the training set into two parts: a training portion and a validation portion [30]. However, Cao and Choe had already separated the images into training and validation sets [22], and therefore we will simply utilize those. After training the model at every epoch, we are able to predict the responses for the validation set which provides us with an estimate of the true generalization error [30]. More importantly, the validation set is employed to tune hyperparameters—variables which we can change in order to control the model's behavior. The model cannot automatically adapt these variables and thus needs to be altered manually. Examples include the model's architecture (e.g. depth), the learning rate, the number of epochs, etc. [32].

In addition to simply employing a validation set, we utilized a method named cross-validation, or more specifically the k-fold cross-validation where k=5. After combining the training and validation sets, we split the data into 5 non-overlapping subsets. For the duration of 5 trials, we would set one of the subsets (a different one each time) to be the validation set while the other 4 become the training set. At the end, the accuracy would be the average accuracy of all 5 trials [32]. Cross-validation allows for low bias, since it is averaging the accuracies of multiple trials, thereby increasing the credibility of our model's results [50].

3.5 Model Considerations

One of the most difficult challenges of machine learning is improving its ability to generalize to external data that it has never seen before [35]. Specifically, one significant problem is overfitting the training dataset. Overfitting results when the trained model becomes more flexible than required, leading to unnecessary levels of complexity which results in a worse performance [51] (see Figure 9). While the model may seem to perform better on the training set, in reality it becomes too adapted towards the training set and stops generalizing towards previously unseen data [52]. We implemented multiple methods in order to reduce this, such as data augmentation, early stopping, batch normalization, dropout regularization, and weight decay.

3.5.1 Data Augmentation

Data augmentation is a method to increase the amount of data while also improving the quality of these datasets. It has been discovered that larger datasets result in better models [54], which is one advantage of data augmentation. Ultimately, data augmentation allows researchers to resolve the issue of overfitting the training set. Data augmentation includes two sections: data warping and oversampling. Data warping means altering existing data while preserving their labels (e.g. geometric transformations, random erasing, etc.) while oversampling means creating synthetic data and adding them to the training set (e.g. mixing images, etc.). The two data augmentation methods could even be combined [35]. For this paper, we utilized data warping through geometric transformations.

### 3.5.2 Early Stopping

Early stopping will terminate the model's training when the model has consecutively underperformed for a fixed number of epochs. More specifically, early stopping checks the validation set for overfitting errors (i.e. when the training error is larger than the validation error by a pre-specified threshold). This prevents the model from getting worse at generalizing to previously unseen data, thereby reducing overfitting [52]. We utilized early stopping for all our models. Furthermore, we set the default patience of early stopping (the number of epochs to pass without improvements on the validation set) to 5 epochs.

### 3.5.3 Batch Normalization

Training deep learning models with stochastic gradient descent (SGD) [33] has been extremely effective, but requires the careful tuning of hyperparameters (i.e. the learning rate and the initial parameter values). In fact, each layer's inputs are affected by the parameters in previous layers, creating cascading effects after small changes. This is problematic as the layers need to constantly adapt to the new distribution, which becomes more difficult as the depth increases. These problems can usually be addressed by utilizing the ReLU activation function, careful initialization, and low learning rates. However, to avoid this problem altogether, we can implement batch normalization. Batch normalization is a method which fixes each layer's input's mean and variance [34]. It does this by normalizing each batch, subtracting the batch's mean and dividing by the batch's standard deviation [35]. Ultimately, this decreases training time, allows us to use higher learning rates without divergence, and also regularizes the model [34].

### 3.5.4 Dropout Regularization

Dropout is a method which randomly drops neurons and all of their connections from the neural network during training (see Figure 10). This prevents the neural network from adapting too much towards its neighboring neurons. Although these co-adaptations may work well for the trained dataset, they also decrease generalization. Dropout will break these co-adaptations by randomly dropping neurons, preventing any neuron from becoming excessively crucial. This in turn reduces overfitting and increases the model's generalization. One side effect, however, is that models that implement the dropout method will typically take longer to train [57].

### 3.5.5 Weight Decay

Another method to increase the model's generalization pertains to weight decay, which prevents the parameter weights for each neuron from growing too large unless it is absolutely necessary. It does this by penalizing the cost function for excessively large weights. In our model, we utilized the L2 regularization, which adds a penalty term to the cost function: lambda times the sum of the weights squared. Lambda is a hyperparameter which controls how firmly we want to penalize large weights. We defaulted lambda to 0.001 as it performed the best on our models. Cortes et al. discovered that the L2 regularization was much more effective than the L1 regularization without decreasing performance which was why we implemented the L2 regularization [58, 59].

# 4. Results

As described in section 3.3.2, the three model architectures presented are AlexNet, VGG16, and a 3-block VGG-style architecture. Since the models utilize binary cross-entropy with a softmax activation function on the output layer, we set the model to decide either 'Flooded/Damaged' or 'Undamaged' based on whichever the function decides has a greater chance (higher than 50%). Furthermore, as described in section 3.3.3, we opted to tune the hyperparameters greedily. In this case, we trained each model based on the same default hyperparameters (few exceptions) and then chose the best to fine-tune the hyperparameters.

Table 2 confirmed that the models were performing well regardless of the training and validation set utilized. Table 3, Table 4, and Table 5 display the different model architectures along with several modifications, and the accompanying accuracies.

In this case, the best model trained (VGG16 with data augmentation, batch normalization, and weight decay) achieved an accuracy of 99.15% (see Table 5 and Figure 11). This means that for every 100 satellite images received, the model will be able to predict about 99 of them correctly. Table 5 also indicates that the model achieves high results on the other values displayed, such as the F1-score. Even on the unbalanced data set, the model achieved an accuracy of 98.81% (see Figure 12).

# 5. Discussion

Interestingly, for AlexNet, implementing batch normalization made the model perform worse on average. In fact, implementing any of the additional modifications (i.e. batch normalization, data augmentation, and weight decay) made the model perform worse on average. This could be due to the inherent dropout layers we already added within the AlexNet model, making the other methods implemented relatively useless. Ultimately, although the AlexNet model performed much worse than the VGGNet architectures, it was much faster to train.

The 3-block VGGNet has the opposite results of AlexNet, where implementing batch normalization made the model perform better on average. Even with additional modifications (i.e. weight decay, dropout regularization, etc.), adding on batch normalization to the model increased its accuracy by 2.85% on average (e.g. adding batch normalization to a model with dropout regularization increased its accuracy by 5.45%). On the other hand, employing data augmentation seemed to worsen the model. Implementing data augmentation, batch normalization, and weight decay simultaneously, however, achieved on average 3.22% higher accuracy than the other runs solely employing data augmentation as a modification. Overall, the VGGNet model achieved 2.93% higher accuracy than AlexNet after the modifications.

The VGG16 architecture had similar results as the 3-block VGGNet, where adding on batch normalization increased the model's accuracy by 0.15% on average, and with data augmentation, batch normalization, and weight decay, the model achieved the highest accuracy of 99.15%. This is a 0.3% increase in accuracy from the 3-block VGGNet model albeit much slower to train.

Compared to Cao and Choe, our approaches were similar due to the large runtimes of the models. However, unlike Cao and Choe's models, we utilized early stopping, weight decay, and batch normalization, which made a difference in our results. Although Cao and Choe did implement dropout and data augmentation, the increased amount of methods utilized could have increased our model's regularization towards overfitting, especially with early stopping. Further, Cao and Choe only tested a few modifications (i.e. CNN, CNN with leaky ReLU activation function, and CNN with dropout and data augmentation). The limited amount of architectures

Cao and Choe tested could have hindered their ability to achieve a better performance. Additionally, we utilized different optimizers. While we implemented stochastic gradient descent, Cao and Choe employed RMSprop and Adam optimizers, which could have also led to differences in accuracy. Overall, our best model performance (VGG16 with data augmentation, batch normalization, and weight decay) achieved 1.86% higher accuracy than Cao and Choe's model for the balanced test set and 1.73% higher accuracy on the unbalanced test set [6].

Recently, more innovative models have been introduced (e.g. ResNet) that we would like to further explore in the future. Furthermore, due to our limited access to a GPU, we restricted the depth of our network. In the future, another network with greater depth could be trained (e.g. VGG19) to compare their performance. Moreover, the models are only trained on data specific to the dataset (the Greater Houston area in the aftermath of Hurricane Harvey [6]). In the future this could be generalized towards other hurricane events, and more data would be extremely useful in hyperparameter tuning and increasing generalization of the model(s). Additionally, due to the gravity of the aftermath, most of these images will not have time to be preprocessed. Thus, it is also crucial for the model(s) to effectively label low quality images (e.g. images partially covered by clouds), further reducing response times [6].

## 6. Conclusion

We conclude that convolutional neural networks possess the ability to automatically and very accurately label buildings as 'Flooded/Damaged' or 'Undamaged'. This paper explored the use of three model architectures (AlexNet, VGG16, and a 3-block VGG-style architecture), with modifications including data augmentation, batch normalization, dropout regularization, weight decay, etc. In the end, our model achieved a top accuracy of 99.15%. However, implementing the most innovative recent models could lead to even better results. Ultimately, CNNs can effectively be employed for flood detection, reducing the necessity for field reconnaissance while also reducing the response time.

**Figures and Tables**

**Table 1**

*Regular 3-block VGG-style architecture*

| Layer Type | Output Shape | Number of Trainable Parameters |
|---|---|---|
| Input | (128, 128, 3) | 0 |
| 2-D Convolutional 32@(3x3)[a] | (128, 128, 32) | 896 |
| 2-D Convolutional 32@(3x3) | (128, 128, 32) | 9,248 |
| 2-D Max pooling (2x2)[b] | (64, 64, 32) | 0 |
| 2-D Convolutional 64@(3x3) | (64, 64, 64) | 18,496 |
| 2-D Convolutional 64@(3x3) | (64, 64, 64) | 36,928 |
| 2-D Max pooling (2x2) | (32, 32, 64) | 0 |
| 2-D Convolutional 128@(3x3) | (32, 32, 128) | 73,856 |
| 2-D Convolutional 128@(3x3) | (32, 32, 128) | 147,584 |
| 2-D Max pooling (2x2) | (16, 16, 128) | 0 |
| Flattening | (32768, ) | 0 |
| Fully Connected | (4096, ) | 134,221,824 |
| Fully Connected | (4096, ) | 16,781,312 |
| Fully Connected | (2, ) | 8,194 |

*Note.* The total number of trainable parameters for this convolutional neural network model is 151,298,338.

[a] A@(3x3) means the convolutional layer has A filters, with each filter having width and height equal to 3. [b] The (2x2) means the pooling layer has width and height equal to 2.

**Table 2**

*Cross Validation for AlexNet and the 3-block VGGNet*

| Measurement | AlexNet | | 3-block VGGNet | |
| :---: | :---: | :---: | :---: | :---: |
| | Without Batch Normalization | With Batch Normalization[a] | Without Batch Normalization | With Batch Normalization[b] |
| Accuracy | 95.43% (0.6430%)[c] | 92.93% (1.678%) | 95.90% (0.8423%) | 96.89% (0.3319%) |
| F1 Score | 0.9543 (0.006895) | 0.9331 (0.01310) | 0.9590 (0.007634) | 0.9688 (0.003515) |
| TPR/Recall | 0.940 (0.0151) | 0.973 (0.0254) | 0.966 (0.0157) | 0.968 (0.00298) |
| TNR/Specificity | 0.969 (0.00840) | 0.885 (0.0570) | 0.952 (0.0286) | 0.970 (0.00538) |
| PPV/Precision | 0.969 (0.00789) | 0.899 (0.0441) | 0.953 (0.0250) | 0.970 (0.00585) |
| NPV | 0.940 (0.0137) | 0.972 (0.0245) | 0.967 (0.0141) | 0.968 (0.00295) |

[a] A batch normalization layer follows every convolutional layer. [b] A batch normalization layer follows every convolutional layer. [c] Number in parenthesis represents the standard deviation.

**Table 3**

*Best Results on Balanced Test Set for AlexNet with Modifications*

| Modification Type | Accuracy | F1 Score | TPR/Recall | TNR/Specificity | PPV/Precision | NPV |
|---|---|---|---|---|---|---|
| Normal | 95.92% | 0.9600 | 0.959 | 0.959 | 0.961 | 0.957 |
| With Batch Normalization | 94.80% | 0.9467 | 0.920 | 0.976 | 0.975 | 0.924 |
| With Batch Normalization and Weight Decay | 94.65% | 0.9481 | 0.978 | 0.915 | 0.920 | 0.977 |
| With Data Augmentation | 93.95% | 0.9390 | 0.934 | 0.945 | 0.944 | 0.935 |
| With Data Augmentation and Batch Normalization | 93.70% | 0.9352 | 0.909 | 0.965 | 0.963 | 0.914 |
| With Data Augmentation and Batch Normalization and Weight Decay | 94.20% | 0.9445 | 0.984 | 0.900 | 0.908 | 0.983 |
| With Data Augmentation and Weight Decay | 90.00% | 0.9025 | 0.924 | 0.876 | 0.882 | 0.920 |
| With Weight Decay | 90.35% | 0.9024 | 0.892 | 0.915 | 0.913 | 0.894 |

**Table 4**

*Best Results on Balanced Test Set for 3-block VGGNet with Modifications*

| Modification Type | Accuracy | F1 Score | TPR/Recall | TNR/Specificity | PPV/Precision | NPV |
|---|---|---|---|---|---|---|
| Normal | 96.75% | 0.9669 | 0.955 | 0.980 | 0.979 | 0.956 |
| With Batch Normalization | 98.60% | 0.9860 | 0.979 | 0.993 | 0.993 | 0.979 |
| With Batch Normalization and Dropout Regularization | 98.75% | 0.9875 | 0.987 | 0.988 | 0.988 | 0.987 |
| With Batch Normalization and Weight Decay | 98.85% | 0.9885 | 0.988 | 0.989 | 0.989 | 0.988 |
| With Data Augmentation | 96.25% | 0.9625 | 0.963 | 0.962 | 0.962 | 0.963 |
| With Data Augmentation and Batch Normalization | 96.05% | 0.9619 | 0.995 | 0.926 | 0.931 | 0.995 |
| With Data Augmentation and Batch Normalization and Dropout Regularization | 96.50% | 0.9641 | 0.935 | 0.995 | 0.995 | 0.939 |
| With Data Augmentation and Batch Normalization and Weight Decay | 98.05% | 0.9805 | 0.974 | 0.987 | 0.987 | 0.974 |
| With Data Augmentation and Weight Decay | 94.40% | 0.9467 | 0.996 | 0.892 | 0.902 | 0.996 |
| With Data Augmentation and Dropout Regularization | 90.95% | 0.9057 | 0.867 | 0.952 | 0.948 | 0.877 |
| With Weight Decay | 98.05% | 0.9804 | 0.973 | 0.988 | 0.988 | 0.973 |
| With Dropout Regularization | 93.30% | 0.9364 | 0.988 | 0.878 | 0.890 | 0.987 |

**Table 5**

*Best Results on Balanced Test Set for VGG16 with Modifications*

| Modification Type | Accuracy | F1 Score | TPR/Recall | TNR/Specificity | PPV/Precision | NPV |
|---|---|---|---|---|---|---|
| With Normal | 98.45% | 0.9845 | 0.988 | 0.981 | 0.981 | 0.988 |
| With Batch Normalization | 98.50% | 0.9850 | 0.982 | 0.988 | 0.988 | 0.982 |
| With Batch Normalization and Dropout Regularization | 98.25% | 0.9825 | 0.977 | 0.988 | 0.988 | 0.977 |
| With Batch Normalization and Weight Decay | 98.70% | 0.9870 | 0.989 | 0.985 | 0.985 | 0.989 |
| With Data Augmentation | 97.45% | 0.9750 | 0.998 | 0.951 | 0.953 | 0.998 |
| With Data Augmentation and Batch Normalization | 95.20% | 0.9505 | 0.921 | 0.983 | 0.982 | 0.926 |
| With Data Augmentation and Batch Normalization and Dropout Regularization | 95.65% | 0.9577 | 0.986 | 0.927 | 0.931 | 0.985 |
| With Data Augmentation and Batch Normalization and Weight Decay* | 99.15% | 0.9915 | 0.991 | 0.992 | 0.992 | 0.991 |
| With Data Augmentation and Weight Decay | 98.45% | 0.9845 | 0.983 | 0.986 | 0.986 | 0.983 |
| With Data Augmentation and Dropout Regularization | 95.20% | 0.9535 | 0.985 | 0.919 | 0.924 | 0.984 |
| With Weight Decay | 98.25% | 0.9825 | 0.981 | 0.984 | 0.984 | 0.981 |
| With Dropout Regularization | 96.75% | 0.9675 | 0.967 | 0.968 | 0.968 | 0.967 |

* Best performing model

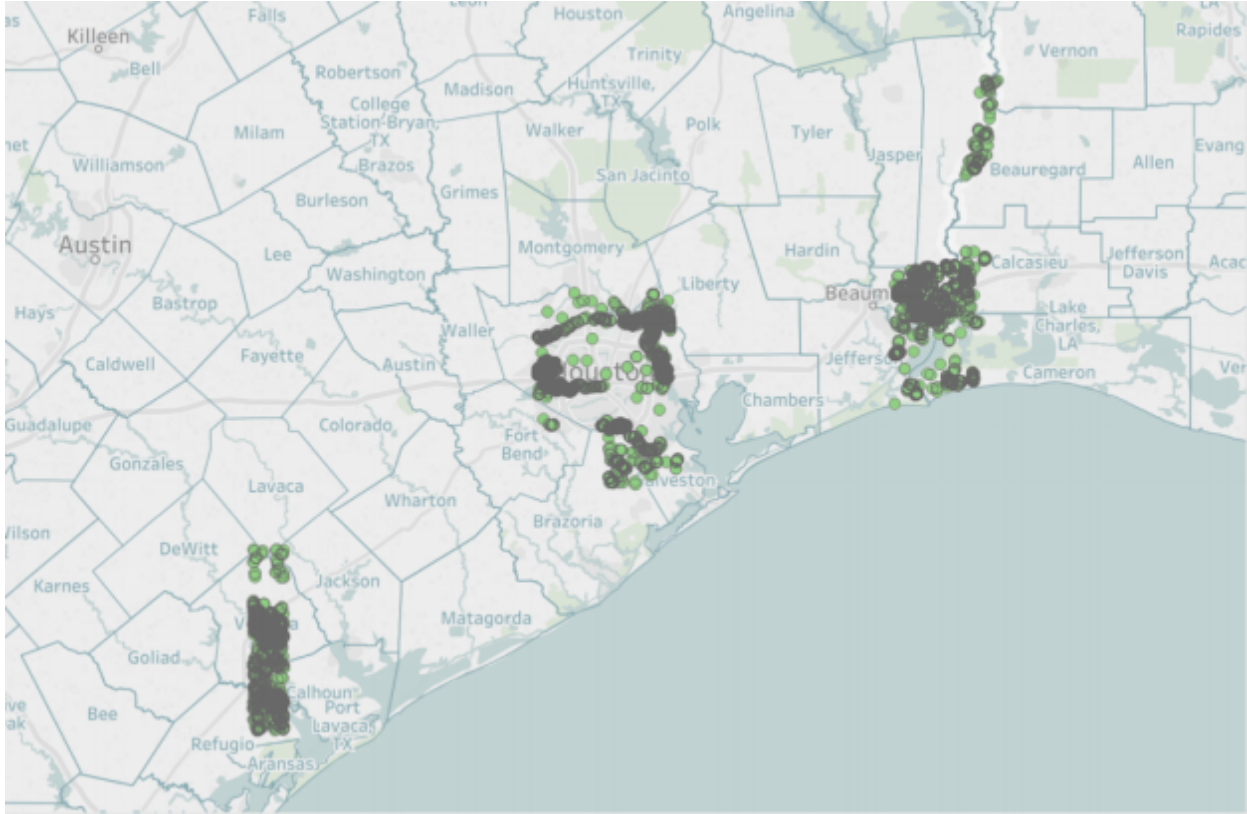

**Figure 1:** Images taken in the Greater Houston area after Hurricane Harvey in 2017, where the green circles represent the regions where the images were taken and labeled [6].

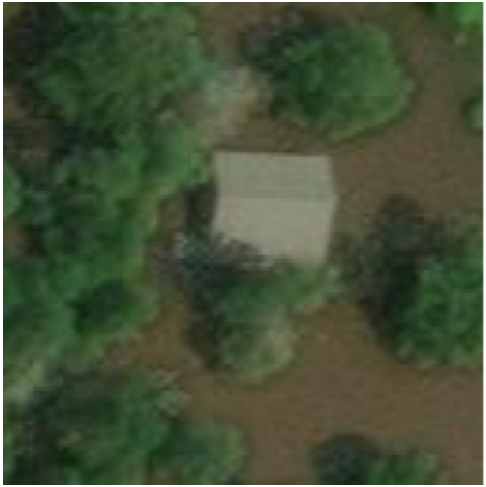 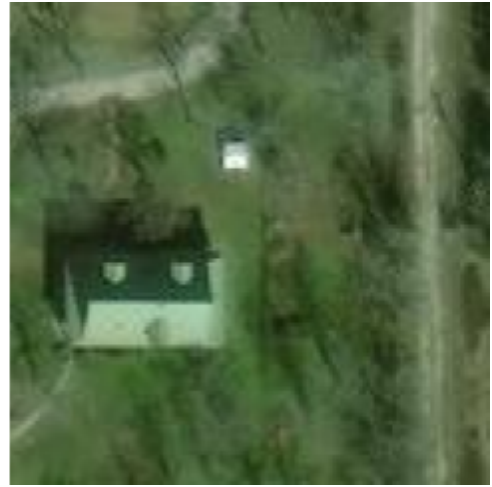

**Figure 2:** Examples of the post-hurricane satellite images. The left image represents a 'Flooded/Damaged' building while the right image displays an 'Undamaged' building [22].

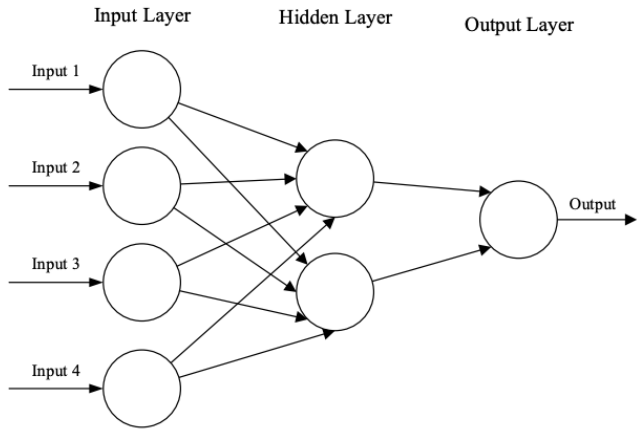

**Figure 3:** Example feedforward artificial neural network with one hidden layer [60].

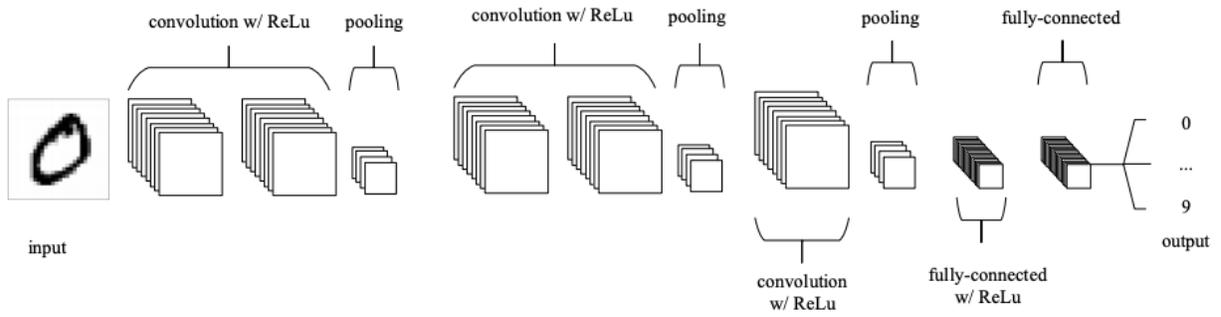

**Figure 4:** Example convolutional neural network architecture for a MNIST dataset [60].

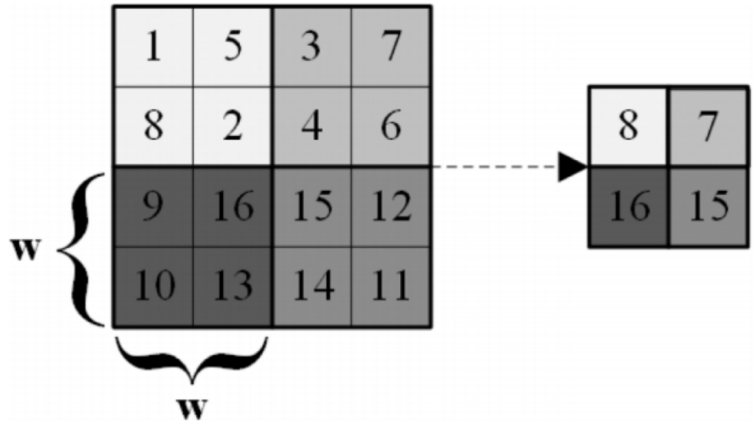

**Figure 5:** Example of a max pooling layer on a net of size 2x2 [37].

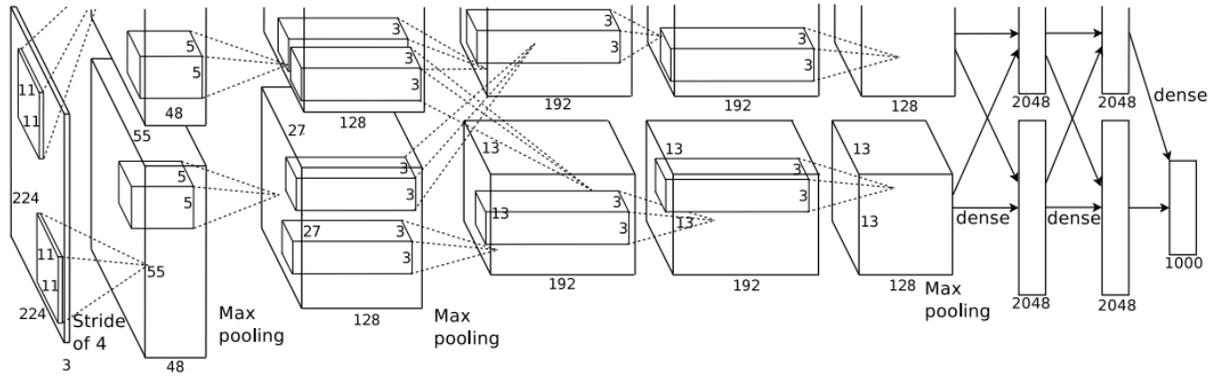

**Figure 6:** AlexNet architecture utilized by Krizhevsky et al. in the ILSVRC in 2012 [23].

| ConvNet Configuration | | | | | |
|---|---|---|---|---|---|
| A | A-LRN | B | C | D | E |
| 11 weight layers | 11 weight layers | 13 weight layers | 16 weight layers | 16 weight layers | 19 weight layers |
| input ($224 \times 224$ RGB image) | | | | | |
| conv3-64 | conv3-64 **LRN** | conv3-64 **conv3-64** | conv3-64 conv3-64 | conv3-64 conv3-64 | conv3-64 conv3-64 |
| maxpool | | | | | |
| conv3-128 | conv3-128 | conv3-128 **conv3-128** | conv3-128 conv3-128 | conv3-128 conv3-128 | conv3-128 conv3-128 |
| maxpool | | | | | |
| conv3-256 conv3-256 | conv3-256 conv3-256 | conv3-256 conv3-256 | conv3-256 conv3-256 **conv1-256** | conv3-256 conv3-256 **conv3-256** | conv3-256 conv3-256 conv3-256 **conv3-256** |
| maxpool | | | | | |
| conv3-512 conv3-512 | conv3-512 conv3-512 | conv3-512 conv3-512 | conv3-512 conv3-512 **conv1-512** | conv3-512 conv3-512 **conv3-512** | conv3-512 conv3-512 conv3-512 **conv3-512** |
| maxpool | | | | | |
| conv3-512 conv3-512 | conv3-512 conv3-512 | conv3-512 conv3-512 | conv3-512 conv3-512 **conv1-512** | conv3-512 conv3-512 **conv3-512** | conv3-512 conv3-512 conv3-512 **conv3-512** |
| maxpool | | | | | |
| FC-4096 | | | | | |
| FC-4096 | | | | | |
| FC-1000 | | | | | |
| soft-max | | | | | |

**Figure 7:** VGG architectures utilized by Simonyan and Zisserman in the ILSVRC in 2014 [23].

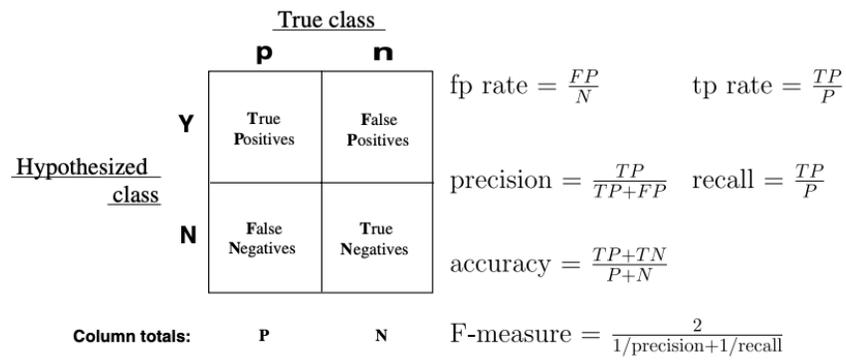

**Figure 8:** Confusion matrix and some common accuracy metric formulas [49].

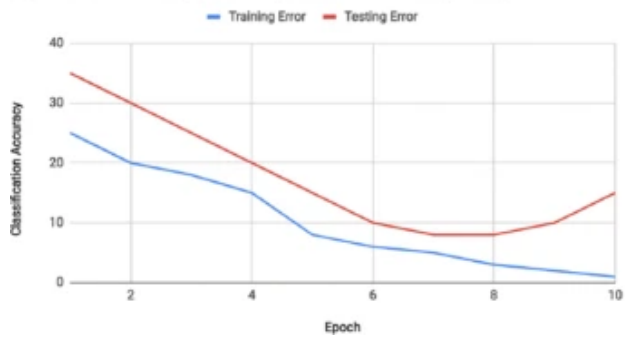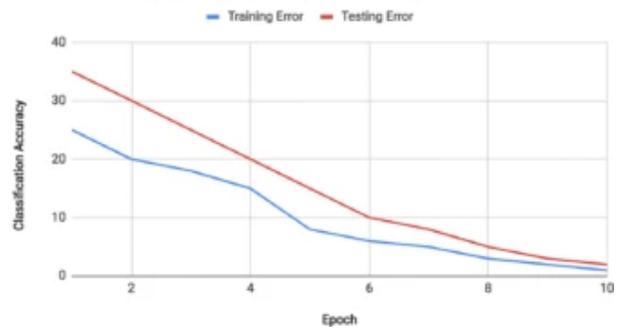

**Figure 9:** The two graphs display the difference between a model who overfits (left) and a model who does not (right). On the left graph, the point where the testing error starts to increase while the training error continues to decrease indicates where the model starts to overfit the training data. Meanwhile, on the right graph, both training and testing errors continue to decrease throughout, which is ideal [53].

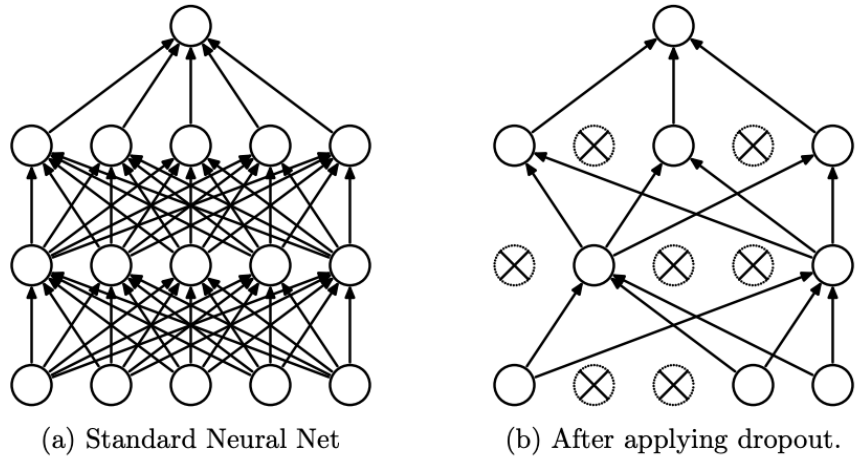

(a) Standard Neural Net  (b) After applying dropout.

**Figure 10:** The image on the left is an example of a standard neural network with two hidden layers. The image on the right is the neural network produced after applying dropout to the network on the left. Crossed out neurons have been dropped out, and connections from crossed out neurons to other neurons have been discarded as well [57].

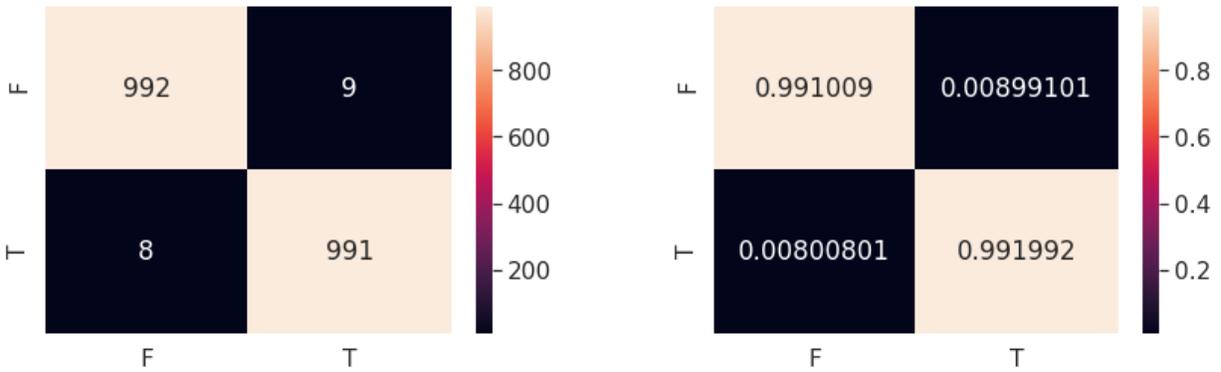

**Figure 11:** These are the confusion matrices for the VGG16 model with data augmentation, batch normalization, and weight decay. These are the results after the trained model encountered the balanced test set, achieving an accuracy of 99.15%. The y-axis displays the true classes, and the x-axis displays the hypothesized classes, where T represents 'Flooded/Damaged' and F represents 'Undamaged'. The left figure represents the confusion matrix while the right represents the normalized confusion matrix.

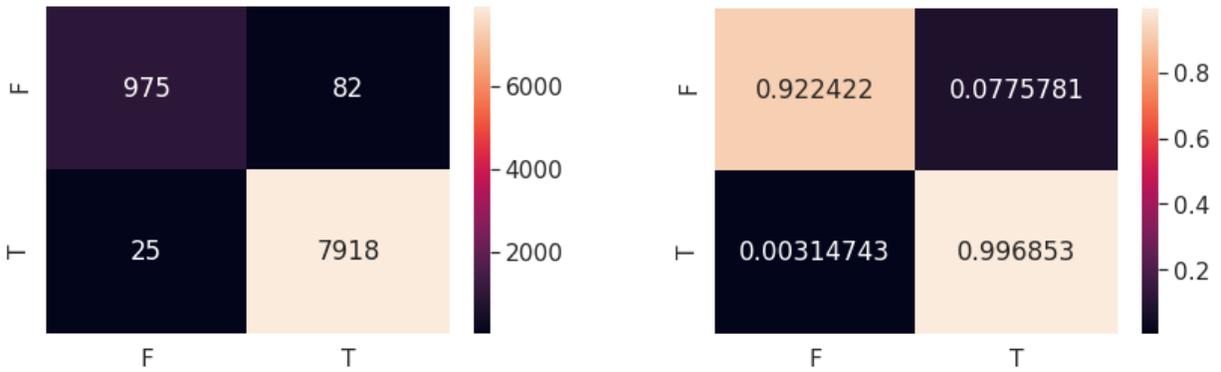

**Figure 12:** These are the confusion matrices for the VGG16 model with data augmentation, batch normalization, and weight decay. These are the results after the trained model encountered the unbalanced test set, achieving an accuracy of 98.81%. The y-axis displays the true classes, and the x-axis displays the hypothesized classes, where T represents 'Flooded/Damaged' and F represents 'Undamaged'. The left figure represents the confusion matrix while the right represents the normalized confusion matrix.